\documentclass[letterpaper, 10 pt, conference]{ieeeconf}  

\IEEEoverridecommandlockouts                              

\usepackage{amsmath,amssymb,amsfonts}
\usepackage{algorithmic}
\usepackage{algorithm}
\usepackage{array}

\usepackage{textcomp}
\usepackage{color}
\usepackage{stfloats}
\usepackage{url}
\usepackage{verbatim}
\usepackage{graphicx}
\usepackage{cite}
\usepackage{balance}
\usepackage{mathrsfs}

\usepackage[colorlinks,
            linkcolor=blue,       
            anchorcolor=blue,  	
            citecolor=blue,        
            ]{hyperref}

\usepackage{graphics} 

\usepackage{amsthm}
\usepackage{multirow}
\usepackage{booktabs}
\usepackage{subfigure} 
\usepackage{float}

\title{\LARGE \bf
Multi-robot Task Allocation and Path Planning with Maximum Range Constraints
}

\author{Gang Xu$^{1}$, Yuchen Wu$^{1, 2}$, Sheng Tao$^{1}$, Yifan Yang$^{1, 2}$, Tao Liu$^{3}$, Tao Huang$^{1}$, Huifeng Wu$^{4}$, \\and Yong Liu$^{1,*}$ 
\thanks{$^{1}$Gang Xu, Yuchen Wu, Sheng Tao, Yifan Yang, Tao Huang, and Yong Liu are with the Institute of Cyber-Systems and Control, Zhejiang University, Hangzhou 310027, China (e-mail: wuuya@zju.edu.cn).}%
\thanks{$^{2}$Yuchen Wu and Yifan Yang are also with the Polytechnic Institute of Zhejiang University, Hangzhou 310015, China.}%
\thanks{$^3$Tao Liu is with the Ocean College, Zhejiang University, Zhoushan 316021, China}
\thanks{$^4$Huifeng Wu is with the Institute of Intelligent and Software Technology, Hangzhou Dianzi University, Hangzhou 310018, China}
\thanks{$^{*}$Yong Liu is the corresponding author (e-mail: yongliu@iipc.zju.edu.cn).}
}

\begin{document}
\maketitle
\thispagestyle{empty}
\pagestyle{empty}

\begin{abstract}
This letter presents a novel multi-robot task allocation and path planning method that considers robots' maximum range constraints in large-sized workspaces, enabling robots to complete the assigned tasks within their range limits. Firstly, we developed a fast path planner to solve global paths efficiently. Subsequently, we propose an innovative auction-based approach that integrates our path planner into the auction phase for reward computation while considering the robots' range limits. This method accounts for extra obstacle-avoiding travel distances rather than ideal straight-line distances, resolving the coupling between task allocation and path planning. Additionally, to avoid redundant computations during iterations, we implemented a lazy auction strategy to speed up the convergence of the task allocation. Finally, we validated the proposed method's effectiveness and application potential through extensive simulation and real-world experiments. The implementation code for our method will be available at https://github.com/wuuya1/RangeTAP.

\end{abstract}



\section{Introduction}

Multi-robot task allocation and path planning are crucial technologies across various applications, such as logistics delivery \cite{chenIntegratedTaskAssignment2021}, power grid inspection \cite{alhassanPowerTransmissionLine2020}, autonomous exploration \cite{zhou2023racer}, and search and rescue operations \cite{duMultiUAVSearchRescue2023}. In these scenarios, a fleet of robots must autonomously allocate tasks and navigate to designated positions to perform the assigned tasks. A key challenge is ensuring that the robots complete all tasks using the shortest possible time or distance cost while adhering to their maximum range constraints. In real-world applications, where robots often have limited endurance, it is essential to account for these constraints in large workspaces. Specifically, the distance a robot moves to perform tasks must not exceed the distance it can travel under its maximum endurance constraint. However, multi-robot task allocation and path planning are known to be NP-hard problems, typically solvable with optimal solutions only in small-scale scenarios. Moreover, robots typically need to avoid environmental obstacles while executing tasks. Therefore, task allocation must consider the additional travel distance from obstacle avoidance behavior, further increasing the difficulty of solving task allocation and path planning.

Several studies have considered and addressed the task allocation and path planning for multiple robots. For instance, Liu et al. \cite{liuIntegratedTaskAllocation2022} integrated task allocation and path planning problems, receiving the real-time in multi-robot systems with a solution approach close to optimal. However, their approach primarily aims at indoor environments, making it unsuitable for large-sized scenarios like urban delivery due to its overload computation. Camisa et al. \cite{camisaMultiRobotPickupDelivery2023} proposed a collaborative task allocation and path planning method for unmanned aerial vehicles (UAVs) and unmanned ground vehicles (UGVs). However, their work did not consider the environmental obstacles. Our previous work \cite{xuDistributedMultiVehicleTask2023} addressed task allocation and motion planning for a fleet of robots in large-sized workspaces with dense obstacles. Nevertheless, this work overlooked the robots' range limits, potentially leading to situations where robots fail to complete their tasks within their endurance limits. Overall, current research rarely tackles the challenge of task allocation and path planning in large-sized scenarios due to overload computation, which often requires considering robots' maximum range limits, as seen in applications such as urban logistics, autonomous exploration, and power grid inspection.

Integrating path planning with task allocation methods is an intuitive solution, where obstacle-avoidance paths from a path planner are employed instead of ideal straight-line paths during task allocation. However, this integration significantly increases the computational complexity of task allocation algorithms, compromising the real-time performance of the multi-robot systems. This challenge arises primarily from the substantial computation time required to solve obstacle avoidance paths using existing global path planners, including search-based methods \cite{hartFormalBasisHeuristic1968, haraborOnlineGraphPruning2011}, sample-based methods \cite{karaman2011sampling, salzman2016asymptotically}, and learning-based methods \cite{wang2020neural, yonetani2021path}. Despite widespread use or focus, these methods are computationally expensive, especially in large-sized maps. 

To address these issues, we developed a fast path planner based on our previous work \cite{xuDistributedMultiVehicleTask2023}, which ensures that paths are close to the shortest possible. In addition, it is noteworthy that auction-based methods, like \cite{choiConsensusBasedDecentralizedAuctions2009,shinSampleGreedyBased2022,liEfficientDecentralizedTask2019}, are commonly employed for task allocation problems due to their efficiency. Therefore, we incorporate the proposed path planner into the auction-based approach \cite{xuDistributedMultiVehicleTask2023}. To speed up the process of task allocation solutions, we further introduced a lazy auction strategy to accelerate the convergence of task allocation algorithms while accounting for the robots' endurance limitations. Finally, we introduce RangeTAP, a novel multi-robot task allocation and path planning method under robots' maximum range constraints. Extensive simulation and real-world experiments validated the proposed pipeline's effectiveness and practicality.

The main contributions are summarized as follows:

\begin{itemize}
  \item We developed a fast global path planner for large-sized maps. Comparisons demonstrated that our planner achieves paths comparable in length to those obtained by the state-of-the-art method while offering approximately an order of magnitude improvement.
  \item We introduced a lazy auction strategy to speed up the task allocation without compromising the quality of the solutions. Meanwhile, we consider the robots' maximum endurance constraints in task assignment, ensuring robots can complete the assigned tasks within their limited endurance.
  \item By integrating the proposed path planner into the task allocation, we proposed a novel method, RangeTAP, to address task allocation and path planning under robots' maximum range constraints. This approach effectively resolves the coupling issue between task allocation and path planning. Finally, we validated the proposed pipeline by extensive simulations and real-world experiments.
\end{itemize}
  

\section{RELATED WORK} \label{RELATEDWORKS}

Multi-robot task allocation methods can be categorized into centralized and decentralized sets, with heuristic-based and auction-based methods being the most prominent in these two categories. In centralized methods, the heuristic methods \cite{weiParticleSwarmOptimization2020,schwarzrockSolvingTaskAllocation2018} can simplify complex problems and have higher search efficiency. In contrast, the auction-based methods \cite{zlotMarketbasedMultirobotCoordination2006,nunesMultiRobotAuctionsAllocation2015} enhance the flexibility and efficiency of task assignment through their dynamic decision-making capability and optimized resource allocation mechanism. However, centralized methods are often impractical for scenarios involving enormous robots due to their high computational complexity. On the other hand, although decentralized methods may not guarantee optimal solutions, their ability to substantially reduce computational complexity highlights their practicality. Among them, the most representative is the auction-based method, like \cite{choiConsensusBasedDecentralizedAuctions2009,shinSampleGreedyBased2022,liEfficientDecentralizedTask2019}, which dynamically iterates the allocation results based on the rewards, significantly reducing the computation time. Recently, the decentralized algorithm proposed in \cite{shorinwaDistributedMultirobotTask2023} has significantly improved the solution quality and computation time by introducing ADMM (Alternating Direction Method of Multipliers) \cite{mateos2010distributed}. However, the works above primarily focus on task allocation. In practice, robots must navigate around obstacles to perform their tasks, which introduces additional travel distances from collision avoidance and affects the quality of task allocation solutions. Therefore, the path planning problem in multi-robot task allocation, a crucial and urgent issue, still needs to be addressed.

Fortunately, many studies have focused on solving task allocation and path planning problems for multiple robots. For instance, Turpin et al. \cite{turpin2014capt} introduced a concurrent task allocation and trajectory planning method that enables robots to combine time-optimized trajectories and speed adjustments. Nawaz et al. \cite{nawazMultiagentMultitargetPath2023} explored the application of Markov Decision Processes (MDP) for addressing the unified problem of task allocation and path planning in underwater environments. The work \cite{nguyenGeneralizedTargetAssignment2019} segmented tasks into groups and utilized conflict-based minimum cost flow algorithms and answer set programming for unified path planning and target allocation. The approach proposed in \cite{aryan2024optimal}  tackled task allocation and path planning issues by combining the advanced SMT solver \cite{de2008z3}  and the improved Conflict-Based Search (CBS) algorithm \cite{zhang2022multi}. Furthermore, Okumura et al. \cite{okumuraSolvingSimultaneousTarget2023} utilized suboptimal algorithms by initializing with an arbitrary target assignment and iteratively applying one-step path planning with target swapping to improve solution efficiency. Despite these advances, ensuring real-time performance for multi-robot systems in large-scale workspaces with rich obstacles remains challenging. The methods proposed in \cite{li2023uav, gao2023amarl} are suitable for solving multi-robot task allocation and path planning in large-scale maps. However, they do not consider environmental obstacles and assume that robots have unlimited endurance, which is unrealistic in practical applications. In our previous work \cite{xuDistributedMultiVehicleTask2023}, although we addressed the issue of obstacles in large-scale environments, we overlooked the coupling of multi-robot task allocation and path planning and also assumed that robots have unlimited endurance. 

Unlike the methods above, the proposed method in this letter addressed the multi-robot task allocation and path planning in large-sized workspaces while considering the robots' maximum range constraints. It ensures that robots can complete their assigned tasks within maximum persistence limits, making the approach more practical than existing methods.

\section{Problem Formulation} \label{ProblemFormulation}

Let $\mathcal{R} = \{1, 2, \ldots, i,\ldots,m\}$ denote the set of robot indices, where $m$ is the number of robots. Each robot must travel to assigned task positions for task performing, with the constraint that each robot can execute at most $L_i$ tasks and cannot exceed its maximum allowable endurance distance $D_{max}^i$ ($i \in \mathcal{R}$). Simultaneously, let $\mathcal{T} = \{1, 2, \ldots,j,\ldots,n\}$ represent the set of task indices, with $n$ being the total number of tasks. Let $\lambda_l$ indicate the travel distance discount factor, primarily to ensure the monotonic decrease of the objective function. In this paper, $\lambda_l$ is set empirically as $0.95$.
 The objective function indicates the total reward value $f_i(\mathcal{T}_i)$ for robot $i$ visiting the tasks in set $\mathcal{T}_i$, which can be expressed as 
\begin{equation}
  \label{Equation1}
  \begin{aligned}
    f_i(\mathcal{T}_i) = \sum_{j=1}^{|\mathcal{T}_i|} \lambda_l^{d(\mathcal{P}_i^j)},
  \end{aligned}
\end{equation}
where $|\cdot|$ denotes the number of elements in a set, $\mathcal{T}_i \subseteq \mathcal{T}$ is the ordered set of tasks allocated to robot $i$, $\mathcal{P}_i$ is the path required for robot $i$ to visit all tasks in $\mathcal{T}_i$, $\mathcal{P}_i^j$ is the sub-path in $\mathcal{P}_i$ from the position of robot $i$ to the position of task $j$, and $d(\mathcal{P}_i^j)$ is the length of the sub-path $\mathcal{P}_i$. Note that the ordered set $\mathcal{T}_i$ indicates that robot $i$ performs the task in $\mathcal{T}_i$ in ascending order of the task indices.

In the task allocation process for robot $i$, the marginal reward $\omega_i(k)$ assigning an unallocated task $k$ to the robot $i$ can be defined as
\begin{equation}
  \label{Equation2}
  \begin{aligned}
    \omega_i(k) = f_i(\mathcal{T}_i \cup \{k\}) - f_i(\mathcal{T}_i), \forall k \in \mathcal{T}, k \notin \mathcal{T}_i.
  \end{aligned}
\end{equation}
By combining Eq. (\ref{Equation1}) with Eq. (\ref{Equation2}), the marginal return $\omega_i(k)$ can be derived as
\begin{equation}
  \label{Equation3}
  \begin{aligned}
    \omega_i(k) = \sum_{j=1}^{|\mathcal{T}_i \cup \{k\}|}  \lambda_l^{d(\mathcal{P}_i^j)} - \sum_{j=1}^{|\mathcal{T}_i|} \lambda_l^{d(\mathcal{P}_i^j)}, \forall k \in \mathcal{T}, k \notin \mathcal{T}_i.
  \end{aligned}
\end{equation}
Note that $\omega_i(k)$ ensures that the objective function is submodular, possessing the critical property of diminishing marginal gains. Specifically, the marginal reward $\omega_i(k)$ for adding a task $k$ will not increase as the number of allocated tasks grows. This property of diminishing marginal returns is crucial as it guarantees the algorithm's convergence, as noted in \cite{choiConsensusBasedDecentralizedAuctions2009}.

Finally, the overall objective function for maximizing the marginal reward across all robots in task allocation can be formulated as
\begin{equation}
  \label{Equation4}
  \begin{aligned}
      & \max \sum_{i=1}^{|\mathcal{R}|} \left( \sum_{j=1}^{|\mathcal{T}_i|} \lambda_l^{d(\mathcal{P}_i^j)} \right) \\
      & \text{s.t.} \quad |\mathcal{T}_i| \leq L_i, \quad \forall i \in \mathcal{R} \\
      & \phantom{\text{s.t.}} \quad \sum_{i=1}^{|\mathcal{R}|} (x_{ij}) = 1, \quad \forall j \in \mathcal{T} \\
      & \phantom{\text{s.t.}} \quad d(\mathcal{P}_i) \leq D_{\max}^i, \quad \forall i \in \mathcal{R},
  \end{aligned}
\end{equation}
where $x_{ij}$ is a binary variable equal to $1$ if task $j$ is assigned to the robot $i$, and $0$ otherwise. The constraints are as follows: 1) the number of tasks allocated to robot $i$ must not exceed its maximum task capacity $L_i$; 2) each task $j$ can be assigned to only one robot, though a robot may be assigned multiple tasks; and 3) the total distance traveled by each robot during task execution must not exceed its maximum range distance $D_{max}^i$. Note that we assume ideal communication among all robots and that they are aware of each other's positions as well as the positions of all tasks.

\section{METHODOLOGY} \label{Methodology}


\subsection{Fast Path Planning}

This part presents our proposed fast path planner, which, unlike traditional search-based methods \cite{hartFormalBasisHeuristic1968, haraborOnlineGraphPruning2011}, does not require a grid map. It enables direct executing path planning in continuous workspaces and is suitable for deployment across maps of various sizes. Our previous work \cite{xuDistributedMultiVehicleTask2023} introduced a lightweight path generation approach using a Guidance Point Strategy (GOS). However, this method is prone to myopic decision-making, potentially leading to local optima. The proposed path planner builds upon the guidance point concept and addresses its limitations.

\begin{figure}[t]
  \centerline{
  \includegraphics[width=.9\linewidth]{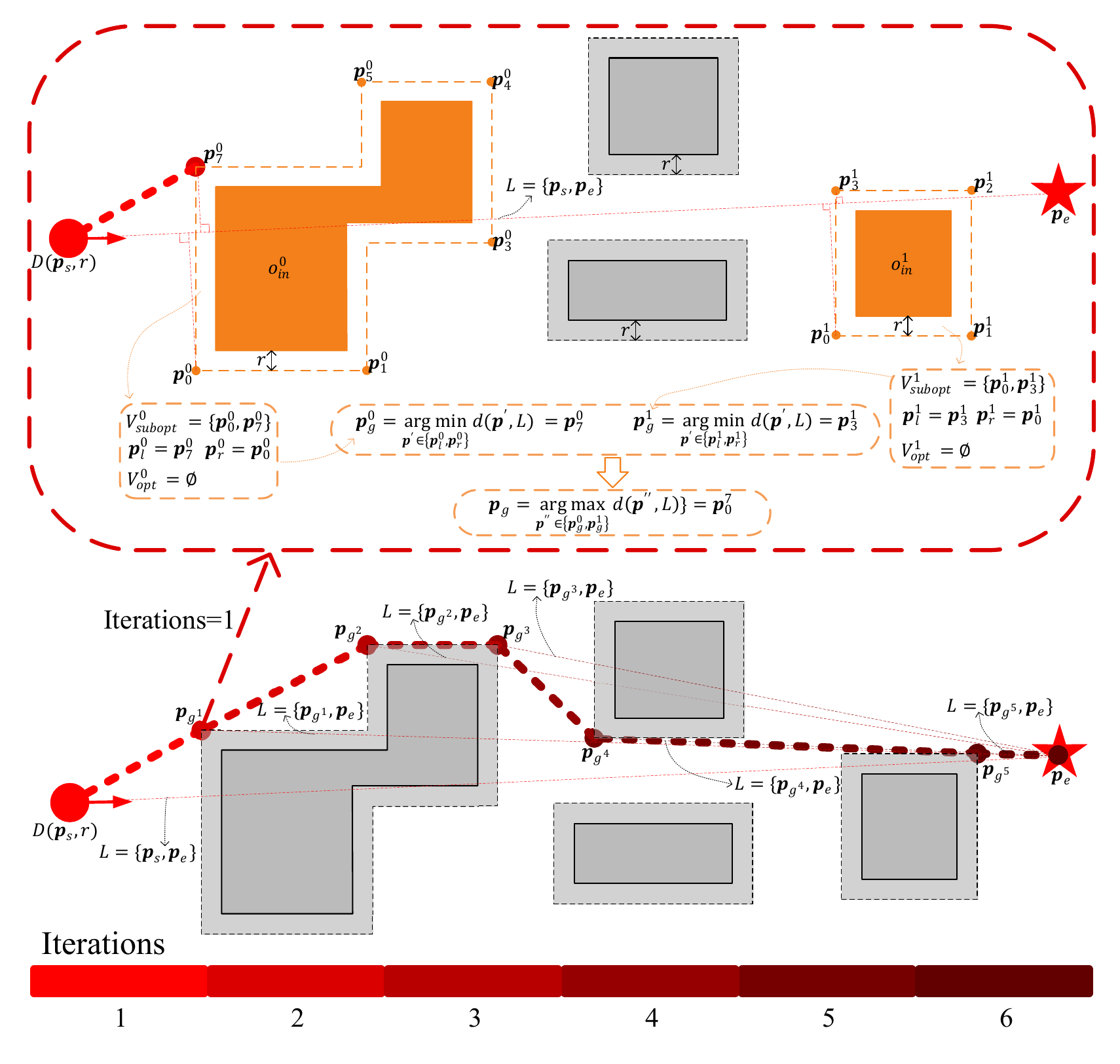}}
  \caption{
    The geometric illustration of the proposed path planner, where $\mathbf{p}_{g^*}$ represents the optimal global guidance point obtained in the *th iteration.
    }
  \label{fig1}
\end{figure}

We first introduce the notations used subsequently. Let the robot's starting and ending points be $\mathbf{p}_s$ and $\mathbf{p}_e$, respectively. Let $\lambda(\mathbf{p}_1,\mathbf{p}_2)$ represent a directed segment in Euclidean space, where $\mathbf{p}_1$ and $\mathbf{p}_2$ are the segment's starting and ending points, respectively, directed from $\mathbf{p}_1$ to $\mathbf{p}_2$. Let the current searched path node be $\mathbf{p}$, with its corresponding guidance point being $\mathbf{p}_g$. Note that at the first iteration, $\mathbf{p}=\mathbf{p}_s$, and $\mathbf{p}_g=\mathbf{p}_e$. To simplify, we regard the robot as a disc with $r$ radius. Let set $O$ indicate all environmental obstacles inflated according to the robot's radius $r$. Note that each inflated obstacle can be approximated by a polygon, allowing us to use polygons to represent various shaped obstacles in the environment, including both convex and concave ones. Additionally, if two inflated obstacles overlap, a polygon can enclose them, treating them as a whole obstacle. It ensures that there is always a traversable path between each inflated obstacle for the robot, guaranteeing the completeness of the algorithm's solution. 
In each iteration, let $\mathbf{p}_g^{opt}$ be the optimal global guidance point, and define $O_{in}$ ($O_{in}\subseteq O$) as the set of all obstacles in the environment that intersect with the segment $L = \lambda(\mathbf{p},\mathbf{p}_g)$. 

Next, we explain how to obtain the optimal global guidance point $\mathbf{p}_g^{opt}$ in each iteration. According to computational geometry, the optimal global guidance point must be among the vertices of the inflated obstacles as long as an obstacle-avoiding global path exists. Assuming $N$ obstacles in $O$ intersect with segment $L$ in a given iteration, we have $O_{in}=\{o_{in}^0,o_{in}^1,\ldots,o_{in}^{N-1},\}$. Let $V_{o_{in}}^i=\{ \mathbf{p}_0^i, \mathbf{p}_1^i, \ldots, \mathbf{p}_{k-1}^i \}$ be the convex vertices set of intersecting obstacle $o_{in}^i$, where $o_{in}^i \in O_{in}$ and $k$ is the number of convex vertices of $o_{in}^i$. Meanwhile, we define another set $V_{subopt}^i$ of convex vertices as
\begin{equation}
  \label{Equation5}
  \begin{aligned}
      V_{\text{subopt}}^{i} = \{\mathbf{p}_{j}^{i} \mid \lambda(\mathbf{p}, \mathbf{p}_{j}^{i}) \cap o_{\text{in}}^{i} = \emptyset\}, \mathbf{p}_{j}^{i} \in V_{o_{in}}^i,
  \end{aligned}
\end{equation}
where the segment $\lambda(\mathbf{p}, \mathbf{p}_j^i)$ formed by any vertex $\mathbf{p}_j^i$ in $V_{\text{subopt}}^i$ and the current searched path node $\mathbf{p}$ does not intersect with obstacle $o_{in}^i$. Further, let $\mathbf{p}_l^i$ and $\mathbf{p}_r^i$ to represent the leftmost and rightmost vertices in $V_{\text{subopt}}^i$ along the direction of segment $L$, respectively. We then define the optimal convex vertices set $V_{\text{opt}}^i$ as
\begin{equation}
  \label{Equation6}
  \begin{aligned}
      V_{\text{opt}}^i = \{\mathbf{p}_v \mid \lambda(\mathbf{p}, \mathbf{p}_v) \cap o_{\text{in}}^i = \emptyset, \lambda(\mathbf{p}_v, \mathbf{p}_g) \cap o_{\text{in}}^i = \emptyset\},
  \end{aligned}
\end{equation}
where $\mathbf{p}_v \in \{\mathbf{p}_l^i, \mathbf{p}_r^i\}$, and both the segment $\lambda(\mathbf{p}, \mathbf{p}_v)$ and $\lambda(\mathbf{p}_v, \mathbf{p}_g)$ do not intersect with obstacle $o_{in}^i$. As illustrated in Fig. \ref{fig1}, two obstacles intersect with segment $L$ in the first iteration. Here, $N=2$, $\mathbf{p}=\mathbf{p}_s$, $\mathbf{p}_g=\mathbf{p}_e$, and $L=\lambda(\mathbf{p}_s,\mathbf{p}_e)$. Taking the intersecting obstacle $o_{\text{in}}^1$ as an example, its corresponding $V_{o_{in}}^1=\{ \mathbf{p}_0^1, \mathbf{p}_1^1, \mathbf{p}_2^1, \mathbf{p}_3^1 \}$, $V_{\text{subopt}}^1=\{\mathbf{p}_0^1, \mathbf{p}_3^1\}$, and the leftmost and rightmost vertices along $L$ being $\mathbf{p}_l^1=\mathbf{p}_3^1$ and $\mathbf{p}_r^1=\mathbf{p}_0^1$, respectively. Meanwhile, according to Eq. (\ref{Equation6}), $V_{\text{opt}}^1 = \emptyset$ for the intersecting obstacle $o_{\text{in}}^1$. 

Thus, we can derive the formulation for the candidate guidance point $\mathbf{p}_g^i$ corresponding to the intersecting obstacle $o_{\text{in}}^i$ as follows:
\begin{equation}
\label{Equation7}
\mathbf{p}_g^i=
\begin{cases} 
\arg \min\limits_{\mathbf{p}' \in V_{\text{opt}}^i} d(\mathbf{p}', L), & \text{if } V_{\text{opt}}^i \neq \emptyset, \\
\arg \min\limits_{\mathbf{p}' \in \{\mathbf{p}_l^i, \mathbf{p}_r^i\}} d(\mathbf{p}', L), & \text{otherwise},
\end{cases}
\end{equation}
where $d(\mathbf{p}’,L)$ represents the distance from vertex $\mathbf{p}’$ to segment $L$. Then, let $P$ be the set of potential optimal global guidance points for all intersecting obstacles, i.e., $P = \{\mathbf{p}_g^0,\mathbf{p}_g^1,\ldots,\mathbf{p}_g^{N-1}\}$. Subsequently, we can deduce the formulation for the optimal global guidance point $\mathbf{p}_g^{\text{opt}}$ as 
\begin{equation}
\label{Equation8}
  \begin{aligned}
    \mathbf{p}_g^{\text{opt}} = \arg \max d(\mathbf{p}'', L), \quad \mathbf{p}''\in P,
  \end{aligned}
\end{equation}
where $d(\mathbf{p}'',L)$ denotes the distance from vertex $\mathbf{p}''$ to segment $L$. As shown in Fig. \ref{fig1}, during the first iteration, the candidate guidance points corresponding to the intersecting obstacles \(o_{\text{in}}^0\) and \(o_{\text{in}}^1\) are \(\mathbf{p}_g^0 = \mathbf{p}_7^0\) and \(\mathbf{p}_g^1 = \mathbf{p}_3^1\), respectively, as determined by Eq. (\ref{Equation7}). At the same time, based on Eq. (\ref{Equation8}), the optimal global guidance point $\mathbf{p}_g^{\text{opt}}$ is $\mathbf{p}_0^7$.

Finally, we summarize the proposed global path planning algorithm into Algorithm \ref{alggo1}. In the first iteration, the robot's current path node $\mathbf{p}$ and the guidance point $\mathbf{p}_g$ are its starting point $\mathbf{p}_s$ and ending point $\mathbf{p}_e$, respectively. Simultaneously, the $\mathbf{p}_s$ is added to its global path $Path$, and the segment $L$ is defined as $L=\lambda(\mathbf{p},\mathbf{p}_g)$. 
Then, all inflated obstacles intersecting with segment $L$ are identified from the set $O$ (see lines 1-3 in Algorithm \ref{alggo1}). After each iteration yields the optimal global guidance point $\mathbf{p}_g^{\text{opt}}$, we add $\mathbf{p}_g^{\text{opt}}$ to the $Path$ if segment $\lambda(\mathbf{p},\mathbf{p}_g^{\text{opt}})$ does not intersect with any obstacles from the set $O$. At the same time, we also update the current path node $\mathbf{p}$, the guidance point $\mathbf{p}_g$, and segment $L$ (see lines 9-14 in Algorithm \ref{alggo1}). If segment $\lambda(\mathbf{p},\mathbf{p}_g^{\text{opt}})$ intersects with any obstacles in set $O$, the guidance point $\mathbf{p}_g$ and segment $L$ will be updated by $\mathbf{p}_g^{\text{opt}}$ (see line 16 in Algorithm \ref{alggo1}). This process is repeated until $\mathbf{p}_g^{\text{opt}}$ is the robot's endpoint $\mathbf{p}_e$ or \(\mathbf{p}_g\) does not exist.
\begin{algorithm}
	\renewcommand{\algorithmicrequire}{\textbf{Input:}}
	\renewcommand{\algorithmicensure}{\textbf{Output:}}
	\caption{Global Guidance Point Strategy (Global-GOS)}
	\label{alggo1}
	\begin{algorithmic}[1]
		\REQUIRE  $\mathbf{p}_s$, $\mathbf{p}_e$, $r$, $O$
		\ENSURE  The global path $Path$ of robot
		\STATE $\mathbf{p} \leftarrow \mathbf{p}_{s}$, $\mathbf{p}_g \leftarrow \mathbf{p}_{e}$
		\STATE $Path \leftarrow Path \cup \{\mathbf{p}_s\}$, ${L} \leftarrow \lambda(\mathbf{p}, \mathbf{p}_g)$
		\STATE $O_{in} \leftarrow CheckIntersect(L, O)$
		\WHILE{}
		\STATE $\mathbf{p}_{g}^{opt} \leftarrow$ Compute the $\mathbf{p}_{g}^{opt}$ with Eq. (\ref{Equation8})
		\IF {$\mathbf{p}_{g}^{opt}$ does not exist}
		\RETURN  $\emptyset$
		\ENDIF 
		\IF {$CheckIntersect(\lambda(\mathbf{p}, \mathbf{p}_{g}^{opt}), O)$ is $\emptyset$}
		\STATE $Path \leftarrow Path \cup \{\mathbf{p}_{g}^{opt}\}$
		\IF {$\mathbf{p}_{g}^{opt}==\mathbf{p}_e$ }
		\RETURN  $Path$
		\ENDIF 
		\STATE $\mathbf{p} \leftarrow \mathbf{p}_{g}^{opt}$, $\mathbf{p}_g \leftarrow \mathbf{p}_{e}$, ${L} \leftarrow \lambda(\mathbf{p}, \mathbf{p}_g)$
		\ELSE 
		\STATE $\mathbf{p}_g \leftarrow \mathbf{p}_{g}^{opt}$, ${L} \leftarrow \lambda(\mathbf{p}, \mathbf{p}_g)$
		\ENDIF 
		\STATE $O_{in} \leftarrow CheckIntersect(L, O)$
		\ENDWHILE
	\end{algorithmic}  
\end{algorithm}

\textit{Complexity Analysis:} As summarized in Algorithm \ref{alggo1}, the Global-GOS planner's computational time complexity primarily arises from searching the guidance point and performing intersection checks. According to the implementation details, the computational complexity of the guidance point search is approximate as \(O(n)\) in the worst case, where \(n\) is the total number of vertices of all polygonal obstacles in the environment. Intersection checks are implemented by determining whether line segments intersect, so the computational complexity is also \(O(n)\) in the worst case. Consequently, in the worst case, the computational complexity for each iteration of the proposed path planner remains approximate as \(O(n)\). Assuming \(k\) iterations, the computational complexity for the planner to find a global path is \(O(k\times n)\). For traditional search-based methods like A* \cite{hartFormalBasisHeuristic1968} and JPS \cite{haraborOnlineGraphPruning2011}, their computational complexity in the worst case is \(O(e + v \log v)\), where $e$ is the number of edges connecting adjacent grid cells, and $v$ is the number of grid cells. In large-sized workspaces, the number of grid cells far exceeds the number of obstacle vertices, resulting in \(O(k\times n)\) $\ll$ \(O(e + v \log v)\). Based on the above analysis, the path planner proposed in this letter is significantly more efficient than traditional search-based methods in large workspaces.

\subsection{Task Allocation Incorporated Path Planning}

This part presents the proposed auction-based method, RangeTAP, for task assignment that incorporates path planning and accounts for the robots' maximum range limits. Traditional auction-based methods typically involve two key phases: the auction phase and the consensus phase. However, these methods generally exploit the ideal straight-line path length for the bids during the auction phase, neglecting the extra travel distance required for obstacle avoidance. It could lead to situations where robots cannot reach their assigned task positions due to limited endurance capabilities. Building on our previous work \cite{xuDistributedMultiVehicleTask2023}, we integrate the proposed Global-GOS planner into the auction phase to obtain the obstacle-avoidance path lengths and ensure assigned tasks are within the robots' endurance limits. Furthermore, we propose a novel lazy auction strategy to speed up the convergence of task allocation by recalculating bids only if they differ from the previous iteration. Therefore, the contribution of task assignment primarily focuses on the auction phase, while the implementation of the consensus phase remains consistent with our previous work \cite{xuDistributedMultiVehicleTask2023}. We describe the implementation details of the auction phase in the Algorithm \ref{alggo}. Readers can refer to \cite{xuDistributedMultiVehicleTask2023} for details on the consensus phase.

The notations used subsequently are consistent with those in Section \ref{ProblemFormulation}. In addition, let $\mathbf{p}_i$ represent the initial position of robot $i$, and $\omega_{cur}^i$ denotes the current reward value for robot $i$ in task allocation, calculated through Eq. (\ref{Equation1}). Let $\sigma_{cur}^i$ be the travel distance required for robot $i$ to visit all task positions in $\mathcal{T}_i$. Furthermore, in each iteration, the marginal reward for assigning task \( j \) to robot \( i \) is denoted as \( \omega_{ij} \). Notably, the marginal reward \( \omega_{ij} \) is the bid that robot \( i \) places for each unallocated task \( j \). In Algorithm \ref{alggo}, the function $\textit{UpdateTA}(\mathcal{T}_i)$ is used to monitor whether a new task has been added to \( \mathcal{T}_i \). The function $\textit{UpdatePath}(\mathcal{T}_i)$ is responsible for obtaining the obstacle-avoidance path for robot $i$ to visit the tasks in \( \mathcal{T}_i \) in sequence. The $d(\mathcal{P}_i)$ as well as $d(Path)$ are used to get the corresponding path length. The function \textit{Global-GOS} is our proposed path planner, with implementation details shown in Algorithm \ref{alggo1}. The notation \(\mathbf{p}(*) \) represents the position of task $*$.

We first describe the proposed lazy auction strategy. In each iteration, we determine whether robot $i$ needs to recalculate the marginal reward \( \omega_{ij} \) through the following two rules: \textit{1)} If robot $i$ has not allocated a new task, its marginal reward \( \omega_{ij} \) for all unallocated tasks $j$ remains unchanged, and there is no need to recalculate. \textit{2)} If robot $i$ is assigned a new task, the marginal reward \( \omega_{ij} \) for all unallocated tasks $j$ should be recalculated. When robot $i$ is assigned a new task, the corresponding increase in path length is added to $\sigma_{cur}^i$, and the corresponding increase in reward is added to the current reward $\omega_{cur}^i$. These steps are illustrated in lines 1-3 of Algorithm \ref{alggo}. Then, we substitute $\omega_{cur}^i$ into Eq. (\ref{Equation3}) to recalculat \( \omega_{ij} \). Since the latter part $\sum_{j=1}^{|\mathcal{T}_i|} \lambda_l^{d(\mathcal{P}_i^j)}$ of Eq. (\ref{Equation3}) always equals $\omega_{cur}^i$ in each calculation, it only needs to be updated once whenever robot \( i \) is assigned a new task. It reduces the computational load during the auction phase, thereby accelerating the convergence of the task allocation process. 

On the other hand, unlike traditional auction phases, our method does not use the ideal straight-line distance to calculate the marginal reward \( \omega_{ij} \). Instead, we use the proposed Global-GOS planner to compute the obstacle-avoidance path length required to assign task \( j \) to robot \( i \). This length is then substituted into Eq. (\ref{Equation3}) to calculate the marginal reward \( \omega_{ij} \). By considering the extra travel distance required for obstacle avoidance, we can more accurately predict whether the travel distance needed for the robot to complete its tasks will exceed its maximum endurance distance. If assigning task \( j \) to robot \( i \) would cause the robot's travel distance to exceed its maximum endurance distance, the corresponding marginal reward \( \omega_{ij} \) is set to sufficiently small but greater than zero. It ensures that the robot can complete its assigned tasks within its maximum range limit. The above process is presented in lines 6-14 of Algorithm \ref{alggo}.





\textit{Analysis:} 
According to the implementation details, the proposed improvement does not increase computational complexity, and the algorithm's optimal approximation guarantee remains consistent with our previous method \cite{xuDistributedMultiVehicleTask2023}. In the worst case, the approximation guarantee is 50\% for monotone submodular objective functions, and the computational complexity is $O(m\times n)$, where $n=\mathcal{|T|}$ is the number of tasks and $m = \mathcal{|R|} \times \mathcal{|T|}$ is the number of task-robot pairs. Readers can refer to \cite{xuDistributedMultiVehicleTask2023} for related details. 


\begin{algorithm}
	\renewcommand{\algorithmicrequire}{\textbf{Input:}}
	\renewcommand{\algorithmicensure}{\textbf{Output:}}
	\caption{Lazy-Based Auctions in RangeTAP}
	\label{alggo}
	\begin{algorithmic}[1]

	\REQUIRE  $j, \omega^i_{cur}, \sigma^i_{cur}, \mathcal{T}_i, D^i_{max}, \mathcal{P}_i, \mathbf{p}_i, r, O$
	\ENSURE  The marginal reward $\omega_{ij}$ of robot $i$ for task $j$
	\IF {$\textit{UpdateTA}(\mathcal{T}_i)$}
	\STATE $\omega^i_{cur} \leftarrow$ Compute the $\omega^i_{cur}$ with Eq. (\ref{Equation1})
	\STATE $\mathcal{P}_i \leftarrow$ $\textit{UpdatePath}(\mathcal{T}_i)$, $\sigma^i_{cur} \leftarrow$ $d(\mathcal{P}_i)$
	\ENDIF 
	\STATE $dist_{temp} \leftarrow \sigma^i_{cur}$
	\IF {$\mathcal{T}_i$ is $\emptyset$ }
	\STATE $Path \leftarrow \textit{Global-GOS}(\mathbf{p}_i, \mathbf{p}(j), r, O)$
	\ELSE 
	\STATE $Path \leftarrow \textit{Global-GOS}(\mathbf{p}(\mathcal{T}_i[-1]), \mathbf{p}(j), r, O)$
	\ENDIF 
	\STATE $dist_{temp} \leftarrow \sigma^i_{cur} + d(Path)$
	\STATE $\omega_{ij} \leftarrow (\omega^i_{cur} + \lambda_l^{dist_{temp}})- \omega^i_{cur}$
	\IF {$dist_{temp} + d(Path)>D^i_{max}$}
	\STATE $\omega_{i} \leftarrow 0.001$
	\ENDIF 	
	\RETURN $\omega_{ij}$

	\end{algorithmic}  
\end{algorithm}

\section{Experiments} \label{Experiments}

In this section, we show the effectiveness of the proposed algorithm through extensive simulations and a real-world experiment. All code is implemented with Python and executed on a 3.2GHz AMD Ryzen 7 7735H Lenovo laptop with 16GB of RAM and RTX 4060 GPU.

\subsection{Ablation Studies} \label{Atitle}

\begin{figure}[thpb]
\centering
\subfigure[$32\,\text{m} \times 32\,\text{m}$]
{
\centering
\includegraphics[scale=0.16]{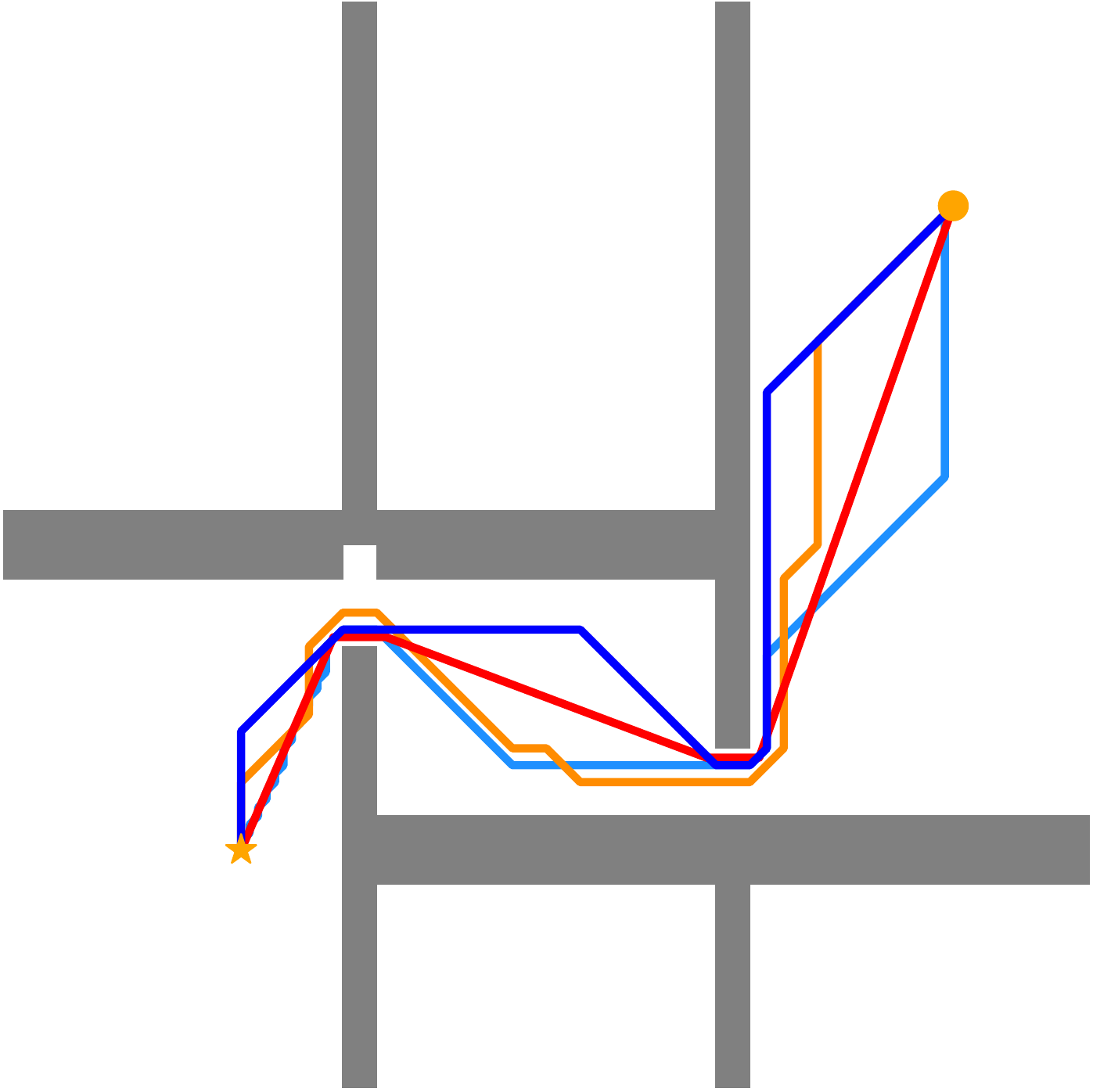}
\label{figsmall}
}%
\subfigure[$256\,\text{m} \times 256\,\text{m}$]
{
\centering
\includegraphics[scale=0.16]{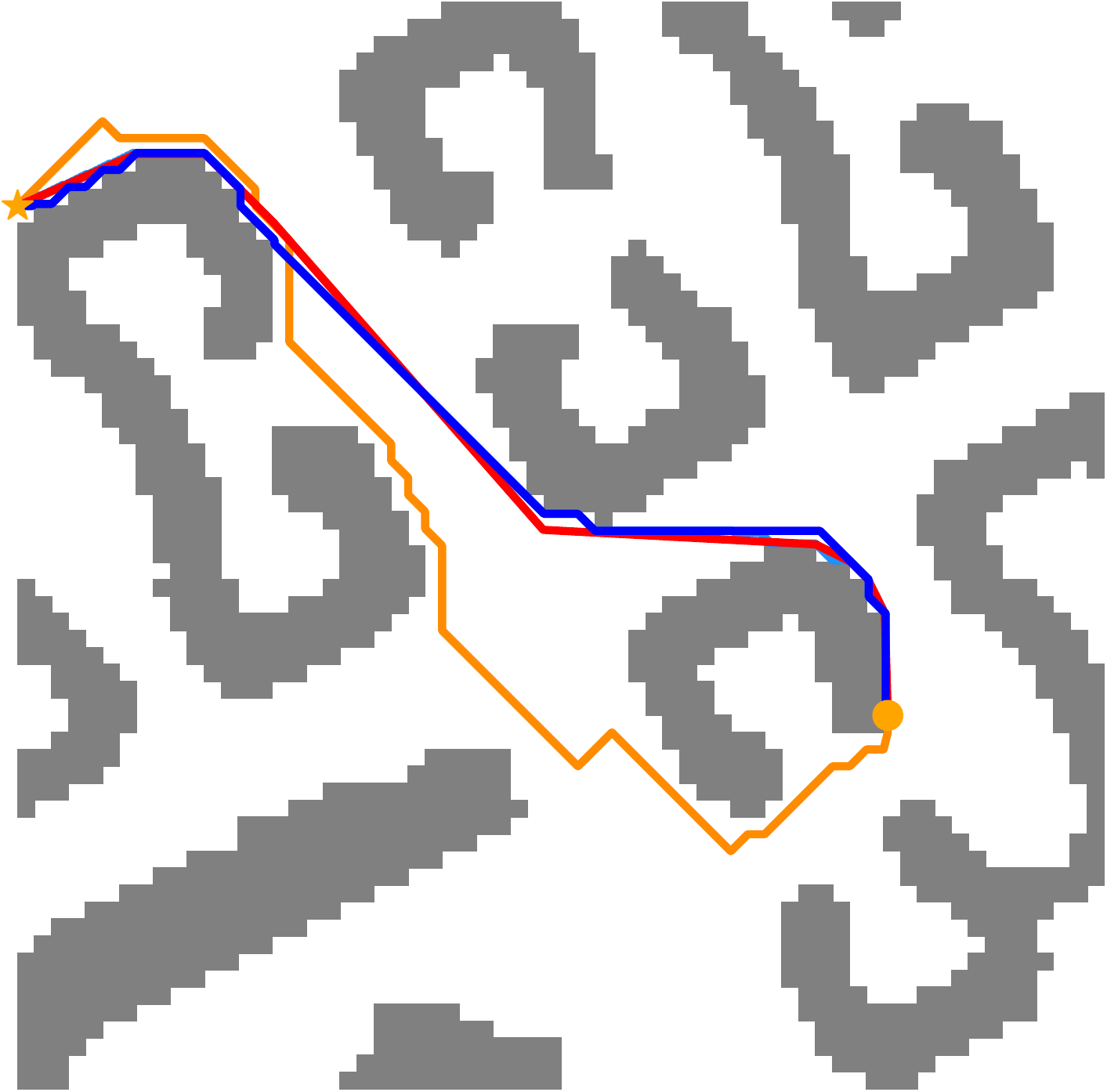}
\label{figmedium}
}%

\subfigure[$6000\,\text{m} \times 4000\,\text{m}$]
{
\centering
\includegraphics[scale=0.252]{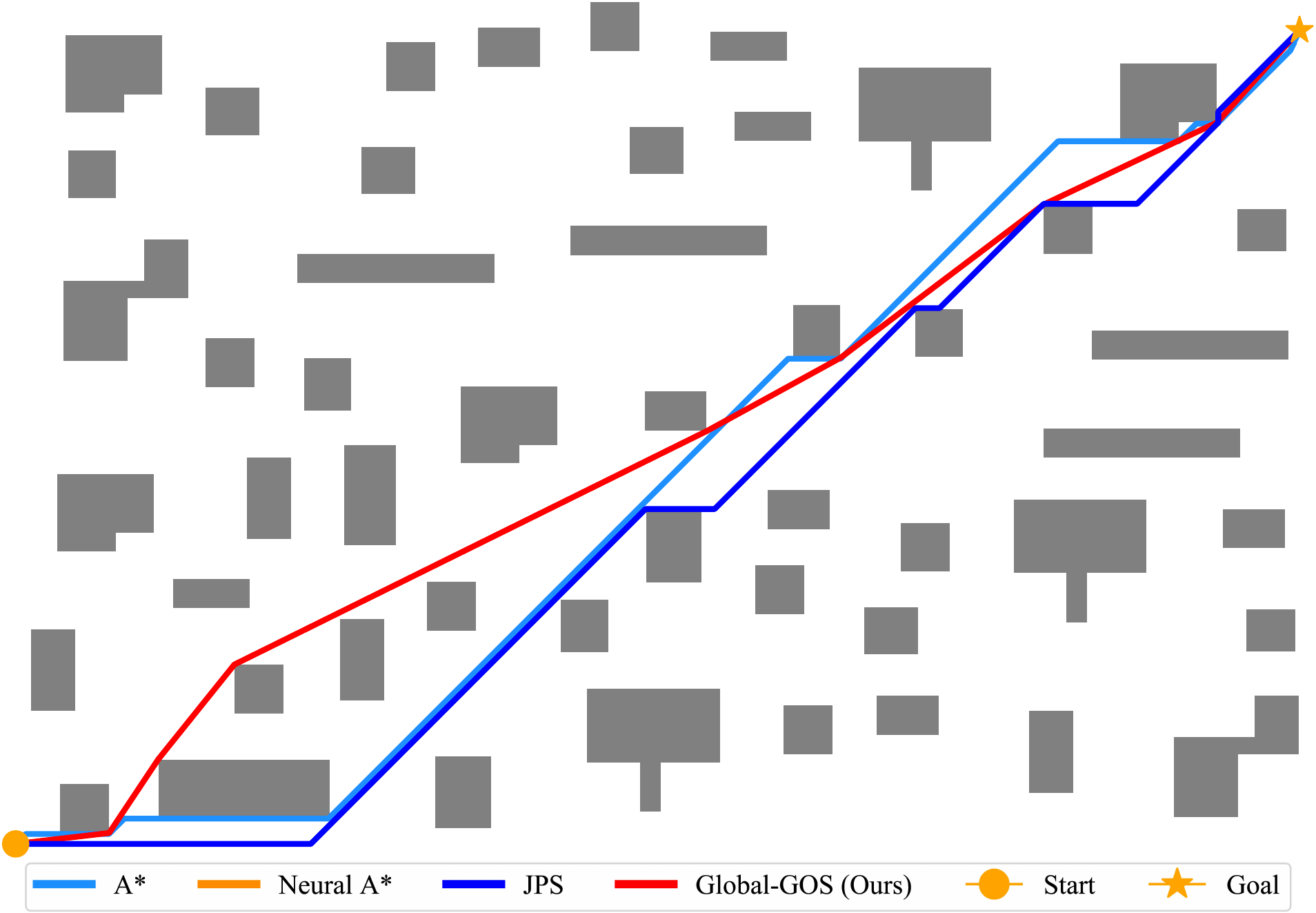}
\label{figlarge}
}%
\centering
\caption{The results of comparisons in path planning. (a) The small-siezd map. (b) The medium-siezd map. (c) The large-siezd map.}
\label{figpathc}
\vspace{-1.0em}
\end{figure}

\begin{table}[h]  
\setlength{\abovecaptionskip}{-0.3cm}
\setlength{\belowcaptionskip}{-0.3cm}
\setlength{\tabcolsep}{5.5pt} 

\centering 
\caption{Compared for Path Planner: \textit{Avg.Time (s)}, \textit{Length (m)}} 
\label{table1}
\begin{center}
\begin{tabular}{cccccccccccccccccccc}    
\toprule Map Sizes & Metrics  &A*  &Neural A*  &JPS  &Ours\\    
\midrule
			Small   
        &\textit{Avg.Time}   &{0.112}  &0.025   &\textbf{0.005}  &\textbf{0.005}\\
		  (32$\times$32)
        &\textit{Length} &40.7   &{41.8}   &{40.4}   &\textbf{37.3}\\ 
\midrule 
        Medium   
        &\textit{Avg.Time}   &{5.197}  &0.222   &0.162  &\textbf{0.029}\\
		  (256$\times$256)
        &\textit{Length} &284.1   &{343.4}   &{284.1}   &\textbf{278.2}\\ 
\midrule 
 		Large   
        &\textit{Avg.Time}   &{11.051}  &-   &4.147  &\textbf{0.041}\\
		  (6000$\times$4000)
        &\textit{Length} &7413.9   &{-}   &{7413.9}   &\textbf{7207.6}\\ 
\bottomrule   
\end{tabular}  
\end{center}
\end{table}

We first compare the proposed Global-GOS planner with three other planners—A* \cite{hartFormalBasisHeuristic1968}, Neural A* \cite{yonetani2021path}, and JPS \cite{haraborOnlineGraphPruning2011}—on maps of three sizes, specifically $32\,\text{m} \times 32\,\text{m}$, $256\,\text{m} \times 256\,\text{m}$, and $6000\,\text{m} \times 4000\,\text{m}$, to demonstrate our planner's efficiency. It is important to note that our planner does not require a grid map for path planning; instead, it treats obstacle regions as polygon-enclosed obstacles. In contrast, the other three algorithms require a grid map in path planning. Each experiment is repeated five times on each map, and the average computation time is recorded to evaluate the performance of the four planners. The results are shown in Fig. \ref{figpathc} and Table \ref{table1}, in which  the \textit{Avg. Time} and \textit{Length} represent the average computation time for path planning and the path length, respectively. Fig. \ref{figpathc} shows that the paths generated by our planner have fewer path nodes than those of the other three planners and are relatively smoother. From Table \ref{table1}, it is evident that our planner outperforms the other planners in terms of average computation time and path length, especially in the large-sized map, where the average computation time is more than 100 times faster than that of the JPS, which is the best-performing among the rest of planners. Note that Neural A* cannot successfully be trained on the large-sized map ($6000\,\text{m} \times 4000\,\text{m}$) due to the excessive number of grids, leading to path planning failure. The experiments demonstrate that our planner is highly effective for quickly predicting the obstacle-avoidance path length in task allocation, ensuring that tasks assigned to robots fall within their maximum range. In addition, the results also show that our planner is promising in ensuring the real-time of robotic systems in large-sized maps.

\subsection{Simulation and Real-world Experiments} \label{Btitle}

\begin{figure}[thpb]
\centering
\subfigure[]
{
\centering
\includegraphics[scale=0.22]{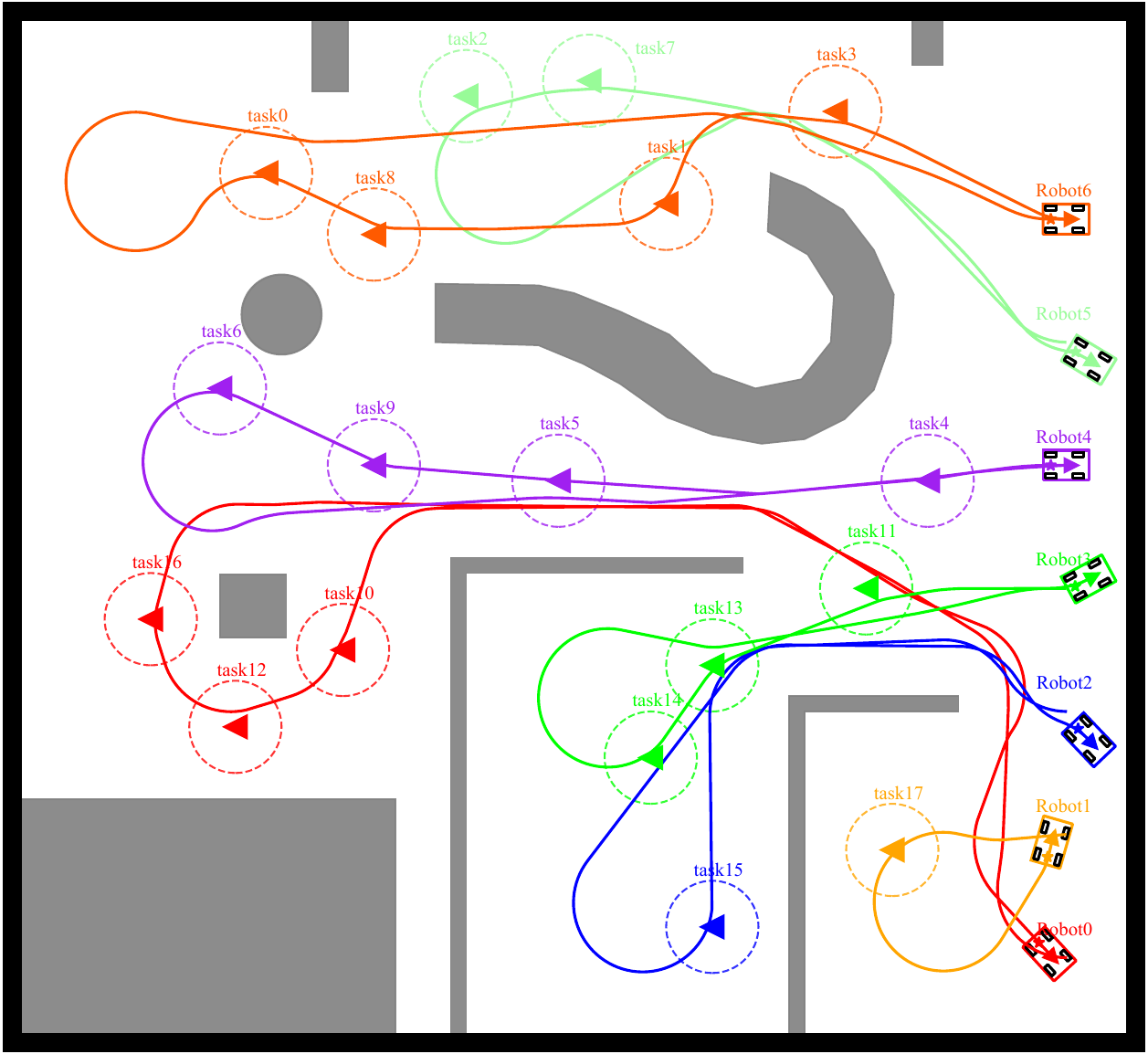}
\label{figstraj}
}%
\subfigure[]
{
\centering
\includegraphics[scale=0.178]{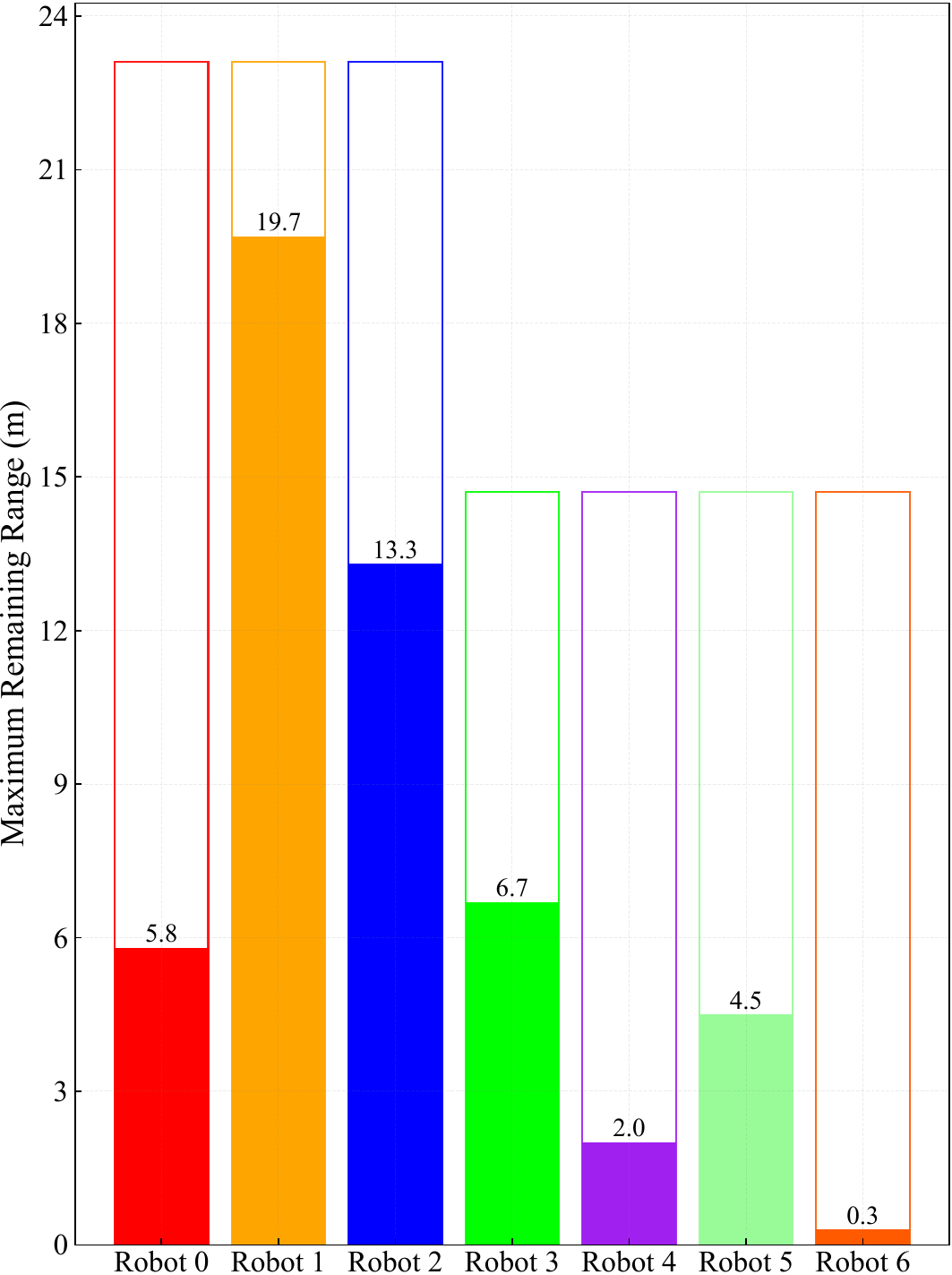}
\label{figsdist}
}%

\subfigure[]
{
\centering
\includegraphics[scale=0.22]{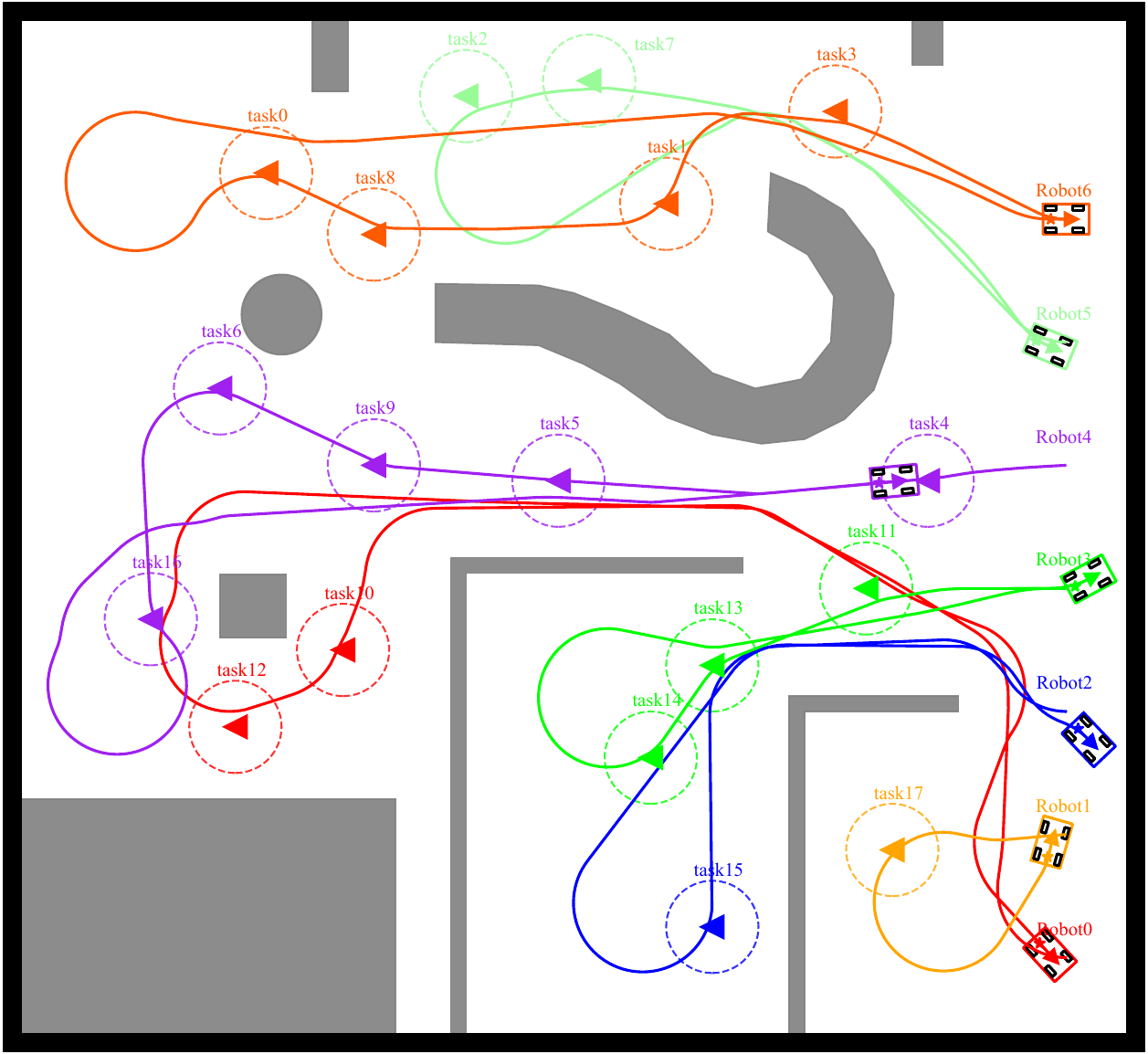}
\label{figstrajc}
}%
\subfigure[]
{
\centering
\includegraphics[scale=0.178]{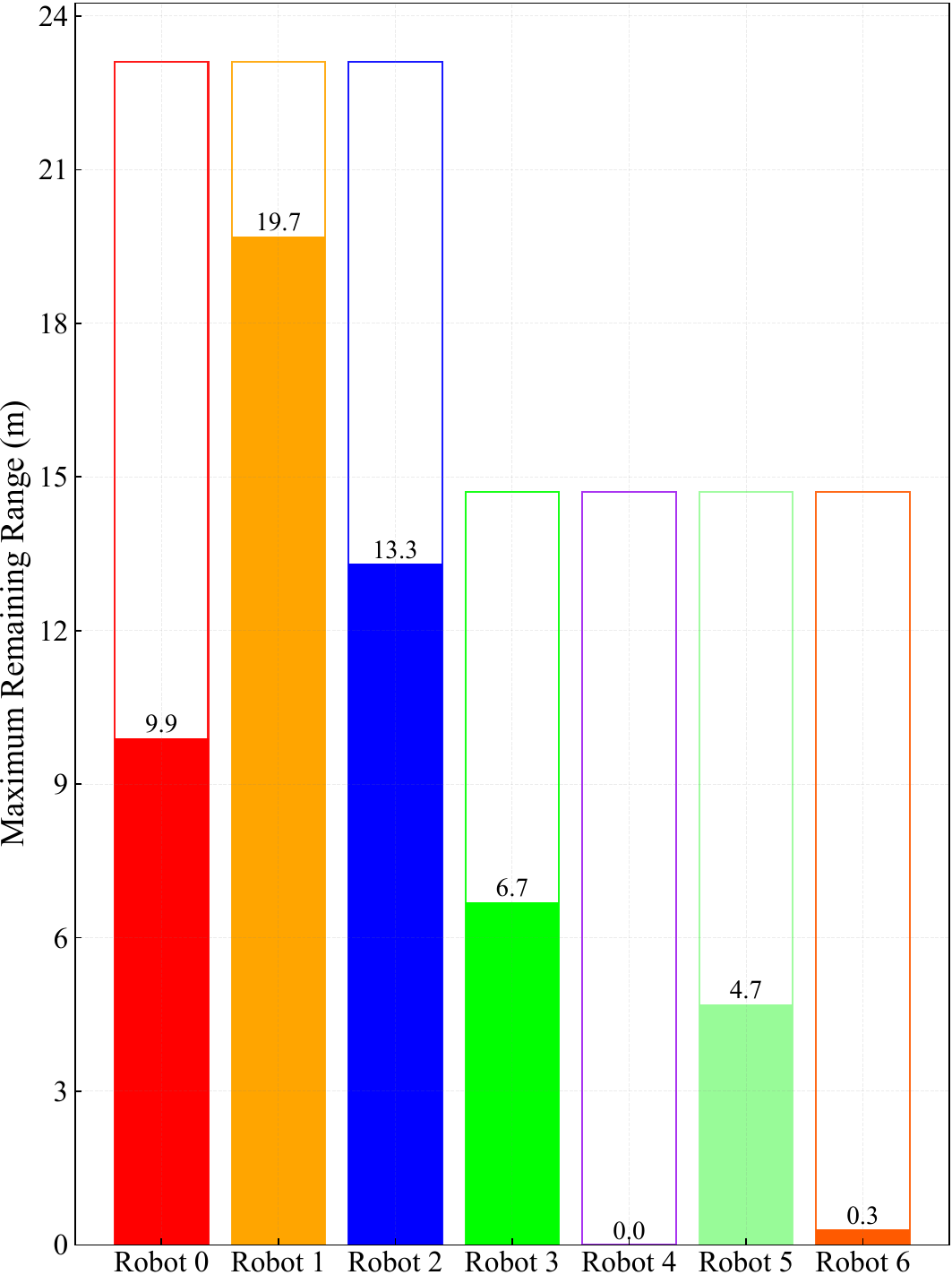}
\label{figsdistc}
}%
\centering
\caption{
Experimental results for task allocation and path planning considering robots' maximum range limits: (a) Trajectories of all robots with the proposed RangeTAP. (b) The maximum remaining range for each robot under the proposed RangeTAP. (c) Trajectories of all robots with the LRGO. (d) The maximum remaining range for each robot under the LRGO.
}
\label{figsim}
\vspace{-1.0em}
\end{figure}

In real-world applications, such as autonomous exploration and logistics delivery, task allocation must consider the robots' maximum range constraints to ensure they complete their assigned tasks within their range limits. Additionally, robots must return to a designated area after completing their tasks. To simulate such scenarios, we randomly deployed 18 task areas and 7 robots in a small, crowded workspace with $6.8\,\text{m} \times 6.6\,\text{m}$, as shown in Fig. \ref{figsim}. We then conducted both simulation and real-world experiments to validate the superiority of the proposed RangeTAP for task allocation and path planning with the robots' maximum range limits. Meanwhile, a task area is believed to be successfully visited when the distance between a robot's position and its task position is less than its radius. The dynamic collision avoidance module uses the improved ORCA algorithm in our previous work \cite{xuDistributedMultiVehicleTask2023}.

We first conduct simulations to compare the proposed RangeTAP with our previous pipeline LRGO \cite{xuDistributedMultiVehicleTask2023}. Here, we selected the Ackermann model unmanned ground vehicles for the experiments. The parameters of the vehicles are set as follows: the radius is $0.18\,m$, maximum speed is $0.12\,m/s$, minimum speed is $0.05\,m/s$, minimum turning radius is $0.45\,m$, and maximum task capacity is $10$. Considering the workspace size, the maximum endurance distance of three vehicles is set as $23.1 \,m$, while the rest of the four is set as  $14.7 \,m$. Finally, the results of the comparisons are shown in Fig. \ref{figsim}. By observing Fig. \ref{figsim}\subref{figstraj} and \subref{figsdist}, it can be found that all robots successfully visit their task area and return to their starting positions by using our RangeTAP. However, when using the LRGO, as shown in Fig. \ref{figsim}\subref{figstrajc} and \subref{figsdistc}, robot 4's remaining range drops to $0\,m$, causing it to stop midway and block robot 0 from proceeding. This occurs because the LRGO does not account for the robots' maximum range constraints during task allocation, leading to task 16 being assigned to robot 4, which exceeded its maximum range.

\begin{figure*}[thpb]
\centering
\subfigure[]
{
\centering
\includegraphics[scale=0.22]{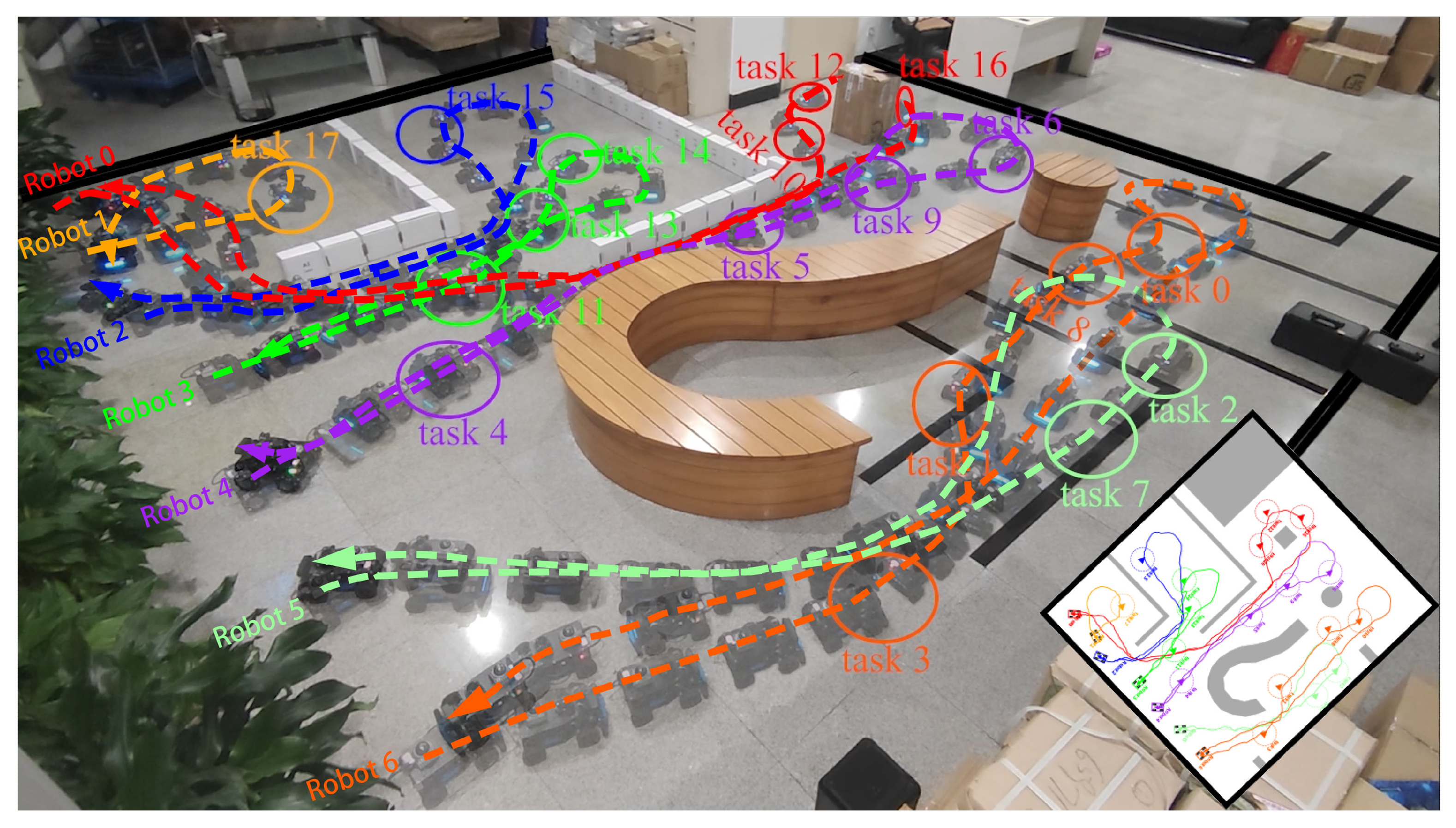}
\label{figrtraj}
}%
\subfigure[]
{
\centering
\includegraphics[scale=0.267]{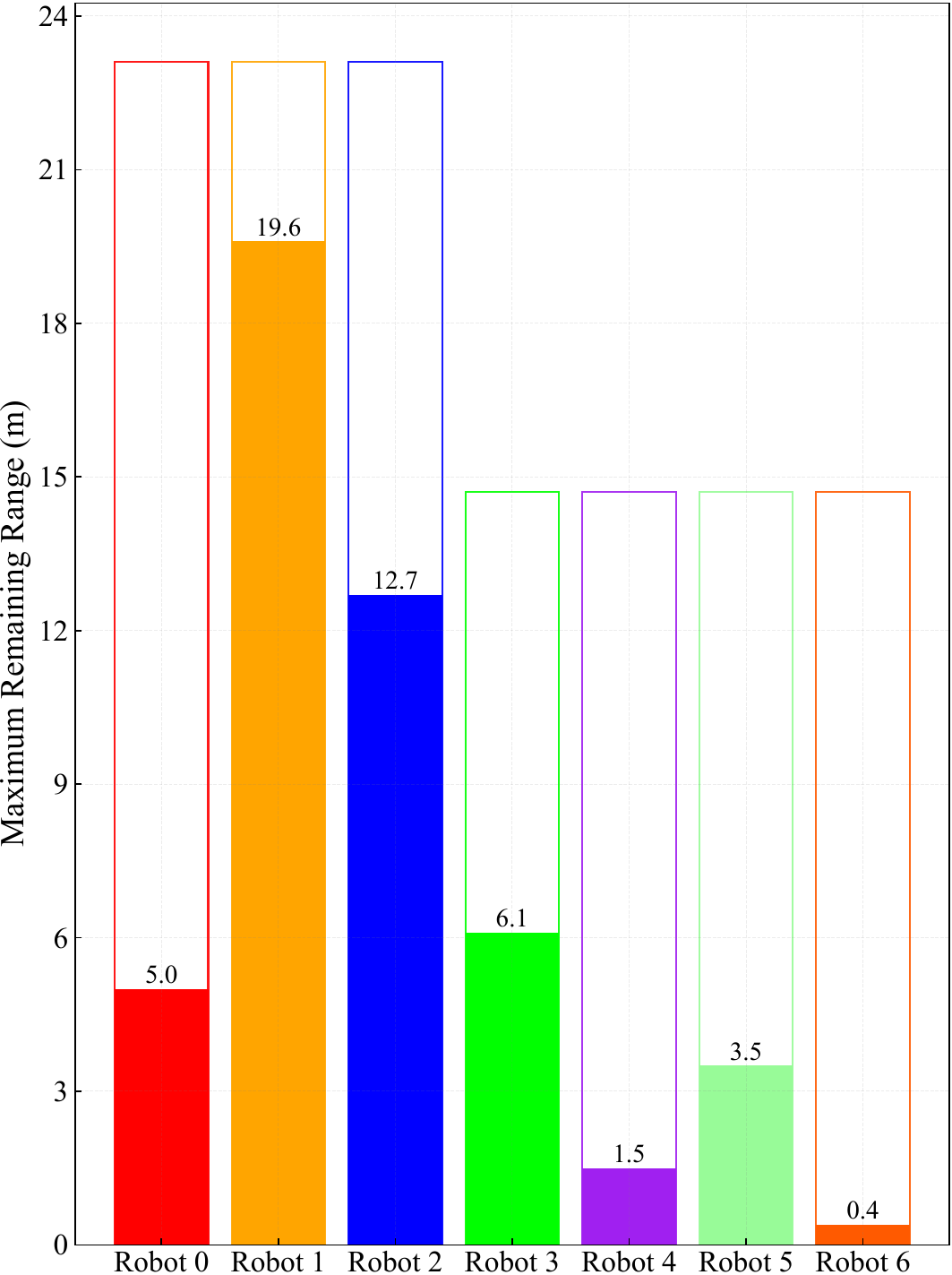}
\label{figrdist}
}%
\centering
\caption{The results of real-world experiment with our proposed RangeTAP: (a) The trajectories of 7 robots visiting 18 task areas. (b) Each robot's maximum remaining range after completing tasks and returning to their respective starting positions. More details can be found in the attached video at https://youtu.be/RY3WLkE3kZs.}
\label{figreal}
\vspace{-1.0em}
\end{figure*}

To further validate the practicality of our proposed pipeline in the real world, we conducted real-world experiment based on the above simulations. All experimental settings are the same as those in the above simulations. Regarding hardware, each robot's computing platform is the NVIDIA Jetson Nano, and we additionally equipped them with Livox Mid360 LiDAR. Meanwhile, we select the Point LIO \cite{he2023point} algorithm as the localization module. Notably, since the LRGO method failed to ensure that each robot successfully returned to its starting position in the above simulations, we only conducted the real-world experiment using our RangeTAP. Fig. \ref{figreal} shows the experimental results. Specifically, Fig. \ref{figreal}\subref{figrtraj} shows the trajectories of 7 robots visiting 18 task areas. Fig. \ref{figreal}\subref{figrdist} shows each robot's maximum remaining range. As shown in Fig. \ref{figreal}, all robots successfully visited their assigned task areas and returned to their starting positions. Additionally, as observed in Fig. \ref{figreal}\subref{figrdist}, each robot still has some remaining range after returning to its starting position. The real-world experiment shows the practicality of the RangeTAP for task allocation and path planning with robots' maximum range constraints.

\subsection{Evaluations in Large-scale Scenarios} \label{Ctitle}

\begin{figure}[t]
  \centerline{
  \includegraphics[width=0.98\linewidth]{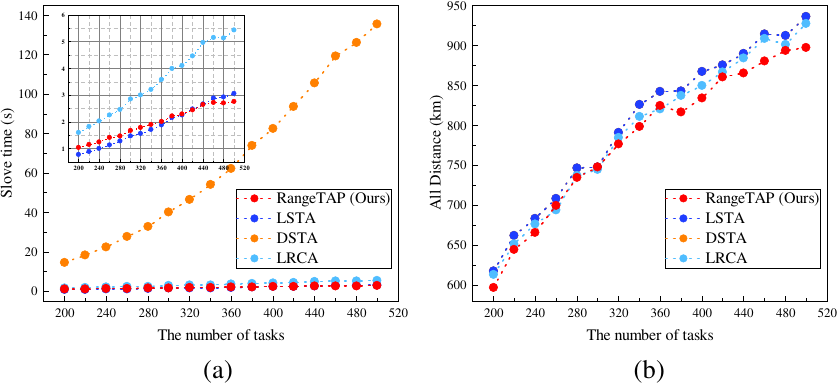}}
  \caption{
    The evaluation results in the large-sized scenario: (a) The average computation time of task allocation for all robot. (b) The average total travel distance required for robots to complete all tasks.
    }
  \label{figexpl}
\end{figure}

This part evaluates the performances of the proposed RangeTAP in the large-sized workspace, as presented in Fig. \ref{figpathc}\subref{figlarge}. We compare the RangeTAP with the DSTA algorithm \cite{shinSampleGreedyBased2022}, LSTA algorithm \cite{liEfficientDecentralizedTask2019}, and the LRCA algorithm proposed in \cite{xuDistributedMultiVehicleTask2023}. Evaluation metrics include the average computation time for all robots to solve task allocation and the average total travel distance required for robots to complete all tasks. For fairness, the other three methods introduced the proposed Global-GOS planner to obtain the obstacle avoidance path length. In the experiments, there were 100 robots and 200 to 500 tasks, and 20 tasks incremented each group test. Each group experiment was repeated ten times for the average computation time and the total travel distance. The positions of all robots and tasks are randomly distributed in the robots' starting area and the workspace, respectively. Additionally, 50 robots have a maximum range of $8000\,m$, while the rest of 50 robots have a maximum range of $20000\,m$. The experimental results are presented in Fig. \ref{figexpl}, where Fig. \ref{figexpl}\textcolor[rgb]{0,0,1}{(a)} represents the average computation time cost for solving task allocation, and Fig. \ref{figexpl}\textcolor[rgb]{0,0,1}{(b)} represents the average total travel distance required for robots to complete tasks. From Fig. \ref{figexpl}\textcolor[rgb]{0,0,1}{(a)}, it is apparent that our proposed RangeTAP achieves the shortest average total distance compared to the other three methods in most cases, while its average computation time is close to the best-performing LSTA algorithm. Significantly, when the number of tasks exceeds 420, our method also demonstrates the shortest average computation time. Moreover, when the number of tasks is 200, our method saves several kilometers of total travel distance compared to the best-performing LRCA, which is significant for saving robots' energy consumption. Overall, our method demonstrates superior performance in computation time and solution quality for task allocation in almost all cases.

\section{Conclusion} \label{Conclusion}

In this letter, we propose a novel multi-robot task allocation and path planning pipeline considering the robots' maximum range constraints. We first introduce a fast path planner to obtain the obstacle avoidance path length in the task assignment phase, in which the experiments show that the proposed path planner is an order of magnitude faster than the current advanced planner. Then, we integrate the proposed path planner into an auction-based task allocation method to overcome the coupling problem between task allocation and path planning. Moreover, we propose a novel lazy auction strategy that saves unnecessary repetitive calculations, accelerating the convergence of the task allocation algorithm without affecting the solution quality. Finally, extensive simulation and real-world experiments show the effectiveness and practicality of our proposed pipeline. In future work, we strive to overcome the dynamic collision problem among robots in dense, dynamic environments while improving the task allocation algorithm and providing theoretical guarantees.

\balance
\bibliographystyle{IEEEtran}
\bibliography{ ours.bib , new_paper_for_iros.bib, mainbibref.bib}

\end{document}